\documentclass{article}

\usepackage[preprint]{neurips_2023}
\usepackage{hyperref}       
\usepackage{parskip}
\usepackage{algorithm}
\usepackage{algpseudocode}
\usepackage{wrapfig}
\usepackage[utf8]{inputenc} 
\usepackage[T1]{fontenc}    
\usepackage{url}            
\usepackage{booktabs}       
\usepackage{amsfonts}       
\usepackage{nicefrac}       
\usepackage{microtype}      
\usepackage{xcolor}         
\usepackage[pdftex]{graphicx}
\usepackage{subcaption}
\usepackage{amsmath}
\usepackage{amsthm}
\usepackage{wrapfig}
\usepackage{authblk}
\newtheorem{theorem}{Theorem}[section]
\newtheorem{proposition}[theorem]{Proposition}

\newtheorem{definition}[theorem]{Definition}
\newtheorem{assumption}[theorem]{Assumption}

\title{Towards understanding neural collapse in supervised contrastive learning with the information bottleneck method}

\author[1]{Siwei Wang}
\author[1,2,3]{Stephanie E Palmer}
\affil[1]{Department of Organismal Biology and Anatomy, University of Chicago}
\affil[2]{Department of Physics, University of Chicago}
\affil[3]{Physics Frontier Center for Living Systems, University of Chicago}
\begin{document}

\maketitle

\begin{abstract}
Neural collapse describes the geometry of activation in the final layer of a deep neural network when it is trained beyond performance plateaus. Open questions include whether neural collapse leads to better generalization and, if so, why and how training beyond the plateau helps. We model neural collapse as an information bottleneck (IB) problem in order to investigate whether such a compact representation exists and discover its connection to generalization. We demonstrate that neural collapse leads to good generalization specifically when it approaches an optimal IB solution of the classification problem. Recent research has shown that two deep neural networks independently trained with the same contrastive loss objective are linearly identifiable, meaning that the resulting representations are equivalent up to a matrix transformation. We leverage linear identifiability to approximate an analytical solution of the IB problem. This approximation demonstrates that when class means exhibit $K$-simplex Equiangular Tight Frame (ETF) behavior (e.g., $K$=10 for CIFAR10 and $K$=100 for CIFAR100), they coincide with the critical phase transitions of the corresponding IB problem. The performance plateau occurs once the optimal solution for the IB problem includes all of these phase transitions. We also show that the resulting $K$-simplex ETF can be packed into a $K$-dimensional Gaussian distribution using supervised contrastive learning with a ResNet50 backbone. This geometry suggests that the $K$-simplex ETF learned by supervised contrastive learning approximates the optimal features for source coding. Hence, there is a direct correspondence between optimal IB solutions and generalization in contrastive learning. Using another backbone pretrained with ImageNet32, we show that a similar compression exists when we perform zero-shot transfer learning on CIFAR10. Training past the plateau in contrastive learning leads to a more generalizable representation.
\end{abstract}

\section{Introduction}
\label{Intro:NC}
Deep neural networks trained for classification exhibit an intriguing geometry, ``neural collapse'' (NC). It is a phenomenon that pervasively occurs in the training paradigm of modern deep neural networks. As a deep neural network is trained beyond the point where the training error plateaus (or reaches zero for specific loss functions; \cite{Papyan2020,han2022neural}), the final layer features collapse to their respective means. This results in classifiers and class means collapsing to the same K-simplex (K is the number of classes) in an equiangular tight frame (ETF) \cite{Strohmer2003}. The emergence of a K-simplex ETF encourages the classifier to classify input by finding their respective nearest class means. This phenomenon was first observed in deep neural networks trained with cross entropy \cite{Papyan2020} and later with mean-square-loss \cite{han2022neural}. Recent work has shown that contrastive-loss-trained models, especially supervised contrastive learning \cite{Khosla2020}, also exhibit neural collapse similar to cross-entropy-trained models \cite{Fang2021}.

However, whether neural collapse leads to good generalization in all cases is unknown. Although \citet{Papyan2020} showed that training beyond zero error improves test accuracy, recent work \cite{Hui2022} has demonstrated that neural collapse may not always occur on the test set. This work also shows that more collapse may impair transfer learning performance. Here, we suggest that when neural collapse corresponds to finding a compact representation of the classification labels $Y$ that maintains lossless prediction \cite{Dubois2021}, neural collapse helps test generalization. To show this, we use the information bottleneck (IB) method to examine and quantify the representations that exhibit neural collapse. IB is a theoretical framework that explicitly trades off input compression with the retention of relevant information. In this case, the relevant information is the label used for classification. In this work, we connect neural collapse with a specific optimal solution of IB, which retains label information by matching the generic compression of the input with label distribution. This solution lends itself to rigorous analytic treatment.

Because the information bottleneck (IB) method and its variations are generally analytically intractable, previous work used variational approximations of IB (VIB \citet{Alemi2016}) to explore how and what deep neural networks encode. Contrastive-loss-trained deep neural networks may be a special case where more precise IB results can be instructive. Recently, advances in nonlinear independent component analysis (ICA) \cite{Hyvarinen2018,Hyvarinen2016} suggested that contrastive loss-trained networks may have an additional desirable property: different instances of the same backbone may converge to the same solution in the representation space. This means that networks trained with this kind of loss may be especially interpretable as they all obtain the same optimal solution. Specifically, \citet{Roeder2020} crystallized this notion and demonstrated that models of the same architecture trained with contrastive loss are linearly identifiable, i.e., their learned representations are equivalent only up to a trivial matrix transformation. This linear identifiability allows us to mobilize the only known closed-form solution of the IB problem, i.e., the Gaussian information bottleneck \cite{Chechik2005} to connect neural collapse in supervised contrastive learning with compression. This extends the previous theoretical prediction in \cite{Papyan2020} that the emergence of neural collapse optimizes feature activations for linear classification to optimality in source coding. 

\textbf{Our contributions}: 
\begin{itemize}
    \item Using the information bottleneck method, we characterize conditions that neural collapse needs to satisify to lead to good test generalization. We suggest that neural collapse corresponds to a special optimal solution for the IB problem of classification, which encodes relevant information $I(Y;Z)$ approximating $H(Y) - \delta $ ($\delta$ is a constant if the performance plateaus or $\delta \rightarrow 0$ if training error goes to zero). We provide two derivations. In the main text, we derive it using the definition of IB. In the Supplementary Information, we include an alternative derivation using an extensioin of rate distortion theory for decoding from \cite{Dubois2021}. 
    
    \item We use the linear identifiability of contrasive learning to approximate the above IB solution. Namely, it allows us to use the Gaussian information bottleneck to find the compact representation $Z$ that retains classification information about $Y$ from $X$. We show that supervised contrastive learning compresses more classification-relevant information into an IB optimal $K$-dimensional Gaussian distribution ($K=10$ for CIFAR10, $K=100$ for CIFAR100) as training progresses. These $K$ dimensions correspond to critical phase transitions in the IB problem of classification. We also show that this $K$-dimensional Gaussian distribution contains the respective $K$-simplex ETF, comparable to the original high-dimensional (2048 in a ResNet50) representation space. These findings suggest that the emerging $K$-simplex ETF from neural collapse is an optimal solution, including all critical phase transitions for the IB problem of classification. Therefore, neural collapse in supervised contrastive learning leads to good test generalization. This also suggests that the corresponding $K$-simplex ETF captures the minimum sufficient statistics from input $X$ about $Y$. Note that our findings suggest that $K$-simplex ETF is optimal for source coding. This is different from the previous optimality result derived in \cite{Papyan2020} which is only about linear classification. 
    
    \item We also show that when we use models pretrained with ImageNet32 as feature extractors, a noisier version of neural collapse occurs in zero-shot transfer learning. A model trained with ImageNet-32 compresses classification relevant information int a $K'$-dimensional Gaussian distribution for CIFAR10 ($K'\sim 70$ for CIFAR10). This suggests that simplex ETF may be a universal geometry employed by modern neural networks reformat classification relevant information.

\end{itemize}
\vspace{-0.1in}
\section{Theoretical Derivation}
\vspace{-0.05in}
\subsection{Notation}
We use $X$ to represent input images and $Y$ to represent their labels. Then, $Y_\textit{train}$ and $Y_\textit{test}$ refer to training labels and test labels, respectively. Throughout the paper, $P(\cdot)$, $Q(\cdot)$, $H(\cdot)$, and $I(\cdot)$ refer to probability distributions, learned probability distributions, entropy, and mutual information, respectively. We also use $Z_{i=1,2}$ to represent learned representations from models (1 and 2) with the same architecture but independently trained. 
To discuss the linear identifiability of supervised contrastive loss, we use $f(\cdot)$ to denote the data representation and $g(\cdot)$ to denote the context representation, respectively. When constructing compressed representations based on IB (i.e., we try to compress $Z_2$ using $Z_1$), we refer to the resulting IB optimal representations as $T$. All datasets $D$ we used here are balanced datasets (equal number of samples for each label). We use $K$ to denote the number of classes in a dataset.
\vspace{-0.1in}
\subsection{Previous work in supervised contrastive learning}
\subsubsection{Supervised contrastive-learning-trained deep neural networks are linearly identifiable}
Given a dataset $D$ with input $x$ and target $y$, a general deep neural network learns an empirical distribution $p_D(y|x)$. Previous work from nonlinear ICA provided sufficient conditions for the learned representation of $p_D(y|x)$ to be linearly identifiable \cite{Hyvarinen2016,Hyvarinen2018}. Namely, given two estimates $\theta$ and $\theta'$ for a data model $P_{D}$, if $P_{D}$ is identifiable, then $p_{D,\theta} = p_{D,\theta'} \rightarrow \theta = \theta'$. Recent work \cite{Roeder2020} extended this notion of linear identifiablity to a broad model family that uses contrastive loss as the objective. Formally, linearly identifiable models can learn representations that are equivalent up to a linear transformation. 
\begin{definition}
If $\theta \stackrel{L}{\sim} \theta'$, then there exists an invertible matrix $M$ such that $q_\theta(x) = M q_\theta'(x)$
\end{definition}
In contrastive-loss-trained neural networks, the loss functions include two representations: a data representation $f_\theta$ and a context representation $g_\theta$. Linear identifiability for contrastive learning indicates that there $\exists M$ and $M'$, $f_\theta(x) = M f_\theta'(x)$ and $g_\theta(x) = M' g_\theta'(x)$. Both $M$ and $M'$ are invertible matrices of rank $K$ when a deep neural network is trained for classification with a $K$-class dataset.  

For the supervised contrastive loss, $z_i$ is the representation of the data $i$, $z_p$ is the representation of other positive samples (sharing the same class label $P(i)$), and $z_a$ is the representation of negative samples in the same minibatch. The loss function can be written as: $p_{\theta}(y|x,S) = \sum_{i\in S} -\log{\left[\frac{1}{P(i)}\sum_{i \in P(i)}\frac{\exp{(z_i \cdot z_p)}}{\sum_{i \in A(i)} (z_i \cdot z_a)}\right]}$, 
where the data representation is
$f(\cdot) = z_i $, context representation is $g(\cdot) = z_p$ and $\sum_{y'\in S}\exp{f_{\theta}(x)^T g_\theta(y')}$ is $\sum_{i \in A(i)} (z_i \cdot z_a)$. Therefore, if we have another model (i.e., $f'(\cdot) = z'_i $, $g'(\cdot) = z'_p$) trained in parallel, there exist matrices $M$ and $M'$ such that $z_i = M\times z'_i$ and $z_p = M'\times z'_p$.

\subsubsection{Neural collapse in supervised contrastive learning} 
\label{section:nc_new}
Neural collapse refers to a phenomenon observed after training error plateaus in a deep neural network whose objective is input classification \cite{Papyan2020,han2022neural}. It is hotly debated whether continuing to train neural networks after the training error goes to zero improves generalization performance \cite{Hui2022}. This phase has been dubbed the ``terminal phase of training'' (TPT) \cite{Ma2017, Belkin2018, Belkin2018a, Belkin2019}. In general, during TPT, the within-class variation of the last-layer features becomes 0. This corresponds to the phenomenon that features collpase to their class means. This is the so-called ``variability collapse'' (NC1 introduced in \cite{Papyan2020}). If the training dataset is balanced, for example there are $n$ samples in each class for all $K$ classes, then the $K$ different class means form an equiangular tight frame (ETF). The ETF formed by the class means has a self-dual property; the optimal classifiers for each class collapse with their class means. As a result, the classification task becomes simply locating the closest class in the feature space for each test sample (NC4 in \cite{Papyan2020}). This makes classification robust against random and adversarial Noise \cite{Papyan2020}. This is the essential element leading to good generalization. 
In contrastive learning, supervised contrastive loss \cite{Khosla2020} is minimized instead of directly optimizing the model's  classification performance. A linear classifier is usually trained independently after the completion of model training \cite{Chen2020,Chen2020a,Khosla2020}. \citet{Fang2021} proved that supervised contrastive learning also exhibits neural collapse (Supplementary Information). Because we only have access to the feature activation $z$, two (NC1 and NC4) neural collapse phenomena discussed above (out of four outlined in \cite{Papyan2020} are applicable (see Supplementary Information for more details).

\vspace{-0.1in}
\subsection{The information bottleneck description of the emergence of K-simplex ETF}
\label{section:nc}
From an information theoretic perspective, training beyond when a deep neural network does not change its training performance means that the mutual information between the training labels and the learned representation remains constant at its maximum once the deep neural network enters TPT, $I(Y_\textit{train};Z) =I_0 = \rightarrow H(Y_\textit{train})-\delta$ ($\delta \rightarrow 0$ as neural collapse occurs). If $I(Y_\textit{train};Y_\textit{test})\rightarrow H(Y_\textit{test})$, the training labels contain sufficient predictive power for the test labels, we will have $I(Y_\textit{test}, Z) \rightarrow H(Y_\textit{test})$.

Combining the above observations, we use $I(Z;Y_\textit{train}) = I_0$ to denote TPT. The constant $I_0$ indicates that the learned representation can obtain no more information about the training labels from the input. 

\begin{equation}
\begin{aligned}
\min_{w,b,\xi} \quad & I(X;Z)\\
\textrm{s.t.} \quad & I(Z;Y_\textit{train}) = H(Y)-\delta\\
\end{aligned}
\label{eq:IB}
\end{equation}

\begin{wrapfigure}{r}{0.5\textwidth}
\includegraphics[width=0.5\textwidth]{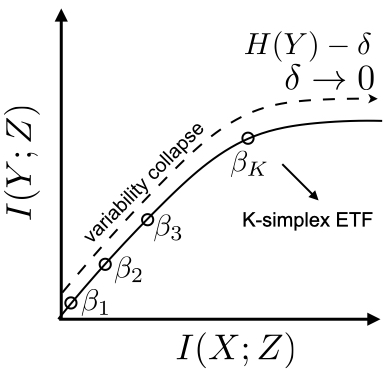}
\vspace{-15pt}
\caption{IB ties compression to better generalization. The IB information curve represents optimal classification across $H(Y)$. We hypothesize that neural collapse exists in the elbow of the IB curve, where the $K$-simplex ETF emerges after the IB optimal solution includes all critical phase transitions $\beta_{1,\cdots,K}$ at equivalent noise levels.}
\vspace{-25pt}
\label{fig:IBcartoon}
\end{wrapfigure}
Our hypothesis is that neural collapse compresses learned representations while improving generalization. This hypothesis is supported by our empirical finding in Section \ref{section:ETF}. In addition, our result in Section \ref{section:collapse} suggests that the compressed representations still approximate the $K$-simplex ETF corresponding to the respective datasets. 
\subsubsection{Approximation of IB objective between linearly identifiable representations using meta-Gaussian Information bottleneck}
\label{theory:MGIB}
Linear identifiability between contrastive learning models implies the existence of a matrix $A$ such that $Z_1 = A \times Z_2$. We propose using a noise model $Z_1 = A \times Z_2 +\xi$ to describe the actual mapping, as linear identifiability is never deterministic. We assume $\xi$ is independent noise with a Gaussian distribution $N(0, \Sigma)$. In the Supplementary Information, we also show that the correlation within class clusters becomes linear as training progresses. Therefore we have the following assumption: 
\begin{assumption}
After the emergence of neural collapse, a Gaussian distribution can approximate the feature activation of all samples belonging to the same class.
\label{assumption1}
\end{assumption}

In Table \ref{table:test_main}, we show that linearly identifiable contrastive learning models achieve comparable generalization performance by correctly predicting the same subset of test samples. This compels us to investigate the structure of the
the mutual information $I(Z_1;Z_2)$. The probability distribution that characterizes $I(Z_1;Z_2)$ is the copula $c(Z_1, Z_2)$ \cite{Ma2008}. A copula is the joint distribution of $U_1$ and $U_2$, where $U_{1,2}$ are rank transformed $Z_{1,2}$, $U_1 = P(Z_1\leq Z^*)$. $U$ is the short notion for cumulative distributions $F(z_1),\cdots, F(z_n)$ (i.e., $F(z_i) = p(z_i \leq z_0)$).  Therefore, $p(Z_1,Z_2) = c(U_1,U_2)\prod_{i=1,2}Z_i$. Formulating $I(Z_1;Z_2)$ with a copula only requires replacing the covariance $Z_{1,2}$ with the copula correlation $corr(U_1, U_2)$. We show the derivation based on \citet{Ma2008,rey2014} in the Supplementary Information. Here, we use $I(Z_1;Z_2) = D_\textit{KL} (p(z_1,z_2)||\prod_{i=1,2} p(z_1)p(z_2))= \int c_{u_1,u_2}\log{c_{(u_1,u_2)}}du_1du_2 = H(c_{u_1, u_2})\\$. Linear identifiability between $Z_1$ and $Z_2$ suggests that $U_1$, $U_2$ are also linearly dependent (both are characterizing the same general correlation structure), $U_1 = A' \times U_2 +\xi$. 

The Assumption \ref{assumption1} suggests that $p(z)$ has a unique Gaussian mixture structure, consisting of $K$ distinct mixture components whose variances are close to zero (thus they have minimal or no overlap). The entropy has a closed form 
$H(z) = \sum_{i=1}^K \left[-\frac{1}{K}\log{\frac{1}{K}} - H(z_i)\right]$. Variability collapse corresponds to $I(Y;Z) \rightarrow -\sum_{i=1}^K\frac{1}{K}\log{\frac{1}{K}} = -\log{\frac{1}{K}} = H(Y)$. As a result, we can use a Meta-Gaussian information bottleneck \cite{Rey2012} to approximate the linearly identifiable portion of $c(Z_1, Z_2)$. Such a Meta-Gaussian information bottleneck (MGIB \cite{rey2014} has computationally tractable optimal solution as the Gaussian information bottleneck \cite{Chechik2005}. IB is a general objective that, in theory, characterizes neural collapse, but its solution is intractable. MGIB and GIB are special cases for which solutions can be calculated. The difference between MGIB and GIB is that MGIB characterizes the compressed representation $T$ from $X$ when $X$ is a Gaussian mixture (with non overlapping mixtures). In this case, the dependence structure between variables are similar between MGIB and GIB, but they differ at their marginals. Linear identifiability in deep neural networks trained with supervised contrastive learning allows us to approximate the intractable IB solution using MGIB/GIB approximation.
Formally speaking, if we define the information bottleneck problem between two learned representations $Z_1$ and $Z_2$ as 
\begin{equation}
\mathcal{L}_{p(t|Z_1), \beta} = I(Z_1;T) - \beta I(Z_2;T)
\end{equation}
Then the following proposition holds when $Z_1 = A \times Z_2 +\xi$. 

\begin{proposition}[Optimality of Meta-Gaussian Information bottleneck]
Consider learned representations $Z_1$ and $Z_2$ with a Gaussian covariance structure and arbitrary margins [\cite{rey2014,Rey2012}]
\begin{equation}
F_{Z_1, Z_2}(z_1,z_2) \sim C_{G}(F_{Z_1}, F_{Z_2})), 
\end{equation} (see Supplementary Information for details)
where $F(Z) = {F_{Z_{1,i}}}$ or ${F_{Z_{2,i}}}$ are the marginal distributions of $Z_1$, $Z_2$ and $C_G$ is a Gaussian copula parameterized by a correlation matrix $G$. The optimum of the minimization problem \ref{eq:IB} is obtained for $T\in \mathcal{T}$, where $\mathcal{T}$ is the set of all random variables $T$ such that $(X,Y,T)$ has a Gaussian copula and $T$ has Gaussian margins. 
\label{prop:MGIB}
\end{proposition}
We provide a sketch of the proof for proposition \ref{prop:MGIB} in the Supplementary Information. 

We use the following theorem from \citet{Chechik2005} to describe the structure of $T$ because the optimal solution $T$ from the Meta Gaussian information bottleneck (MGIB) is also the optimal solution for the corresponding Gaussian information bottleneck. $U$ is the short notion for cumulative distributions $F(z_1),\cdots, F(z_n)$. $\Phi(\cdot)$ is the univariate Gaussian quantile function applied to each component,  $\Phi^{-1}(u) = (\Phi^{-1}(u_1), \cdots, \Phi^{-1}(u_n))$.

\begin{theorem}[Optimal solution for the Gaussian Information Bottleneck]
The optimal projection $T = A'\times U_1+\xi$ for a given tradeoff parameter $\beta$ is given by $\xi = I_x$ and 
\begin{equation}
  A =
    \begin{cases}
      [0^T;\cdots;0^T] & 0\leq\beta\leq\beta_1^c\\
      [\alpha_1v_1^T, 0^T; \cdots ; 0^T] & \beta_1^c\leq\beta\leq\beta_2^c\\
      [\alpha_1v_1^T, \alpha_2v_2^T,0^T; \cdots ; 0^T] & \beta_2^c\leq\beta\leq\beta_3^c\\
      \vdots & \text{otherwise},
    \end{cases}       
\end{equation}
where $v_1^T, \cdots, v_n^T$ are left eigenvectors of $\Sigma_{x|y}\Sigma_x^-1$ sorted by their corresponding ascending eigenvalues $\lambda_1, \cdots, \lambda_n$, $\beta_i^c = \frac{1}{1-\lambda_i}$ are critical $\beta$ values, $\alpha_i$ are coefficients defined by $\alpha_i = \sqrt{\frac{\beta(1-\lambda_i)-1}{\lambda_i r_i}}$, 
$r_i = v_i^T\Sigma_x v_i$, $0^T$ is an n-dimensional row vector of zeros, and semicolons separate rows in the matrix $A$. 
\end{theorem}
$\alpha_i$ emphasizes the difference between consecutive $\lambda$ instead of a single $\lambda$. Each $\alpha_i$ indicates, from an information-theoretic perspective, how much relevant information a given eigenvector contributes to the compressed representation $T$. In Fig. \ref{fig:IBalpha_cifar10} (Section \ref{section:imagenet}), we show that the optimal representation for CIFAR10 has each of the 10 eigenvectors contributing nearly equally. This geometry is similar to that desired for a $K$-simple ETF. 

\vspace{-0.1in}
\section{Experimental Verification}
\vspace{-0.05in}
We first show that as neural collapse happens, the linear identifiability between learned representations for datasets with few classes gets better (Fig. \ref{fig:NC}). This lets us use the meta-Gaussian information bottleneck (which we introduced in Section \ref{theory:MGIB}) between the representations learned by two independently trained ResNet50 backbones to approximate the IB optimal compression solution. For datasets with a small number of classes (CIFAR10 and CIFAR100), we find that the IB optimal representations compress the majority of the classification information into a $K$-dimensional representation ($K$ is the number of classes; for CIFAR10, K=10). In addition, we observe that as generalization (measured by test accuracy) improves, more classification-relevant information is compressed into this $K$-dimensional representation (in Supplementary Information). This ties improved generalization to greater compression. While we do not observe neural collapse in deep neural networks trained on the entire ImageNet32 dataset, we find that when we use these deep neural networks as feature extractors for zero-shot transfer learning, neural collapse also occurs. Using CIFAR10 as an exmple, we find that zero-shot transfer learning packs the classification information into a $K'$-dimensional representation with ($K'>K$, K=10 for CIFAR10), but both the $K$-dimensional and $K'$ dimensional representations share similar geometry (Fig. \ref{fig:IBzeroshot}). This suggests that the simplex ETF may be a universal feature geometry that modern deep neural networks achieve for obtaining efficient representations of the input.
\vspace{-0.1in}
\subsection{Linear identifiability improves \textit{after} the emergence of neural collapse}
\label{result:train}
\begin{figure}
     \centering
     \begin{subfigure}[b]{0.4\textwidth}
     \caption{}
         \centering
         \includegraphics[width=0.6\textwidth]{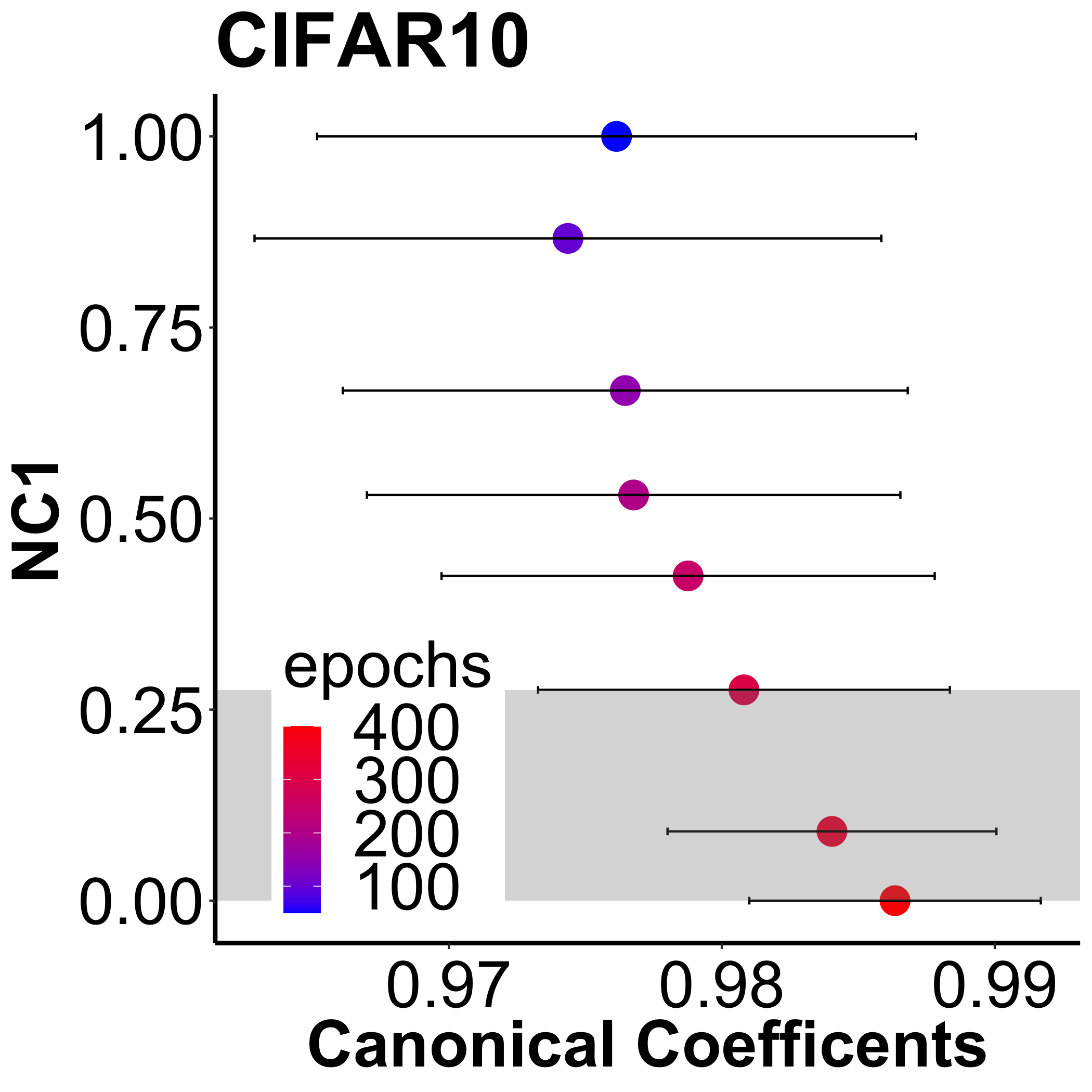}
     \label{fig:eequinorm}
\end{subfigure}
     \begin{subfigure}[b]{0.4\textwidth}
     \caption{}
         \centering
         \includegraphics[width=0.6\textwidth]{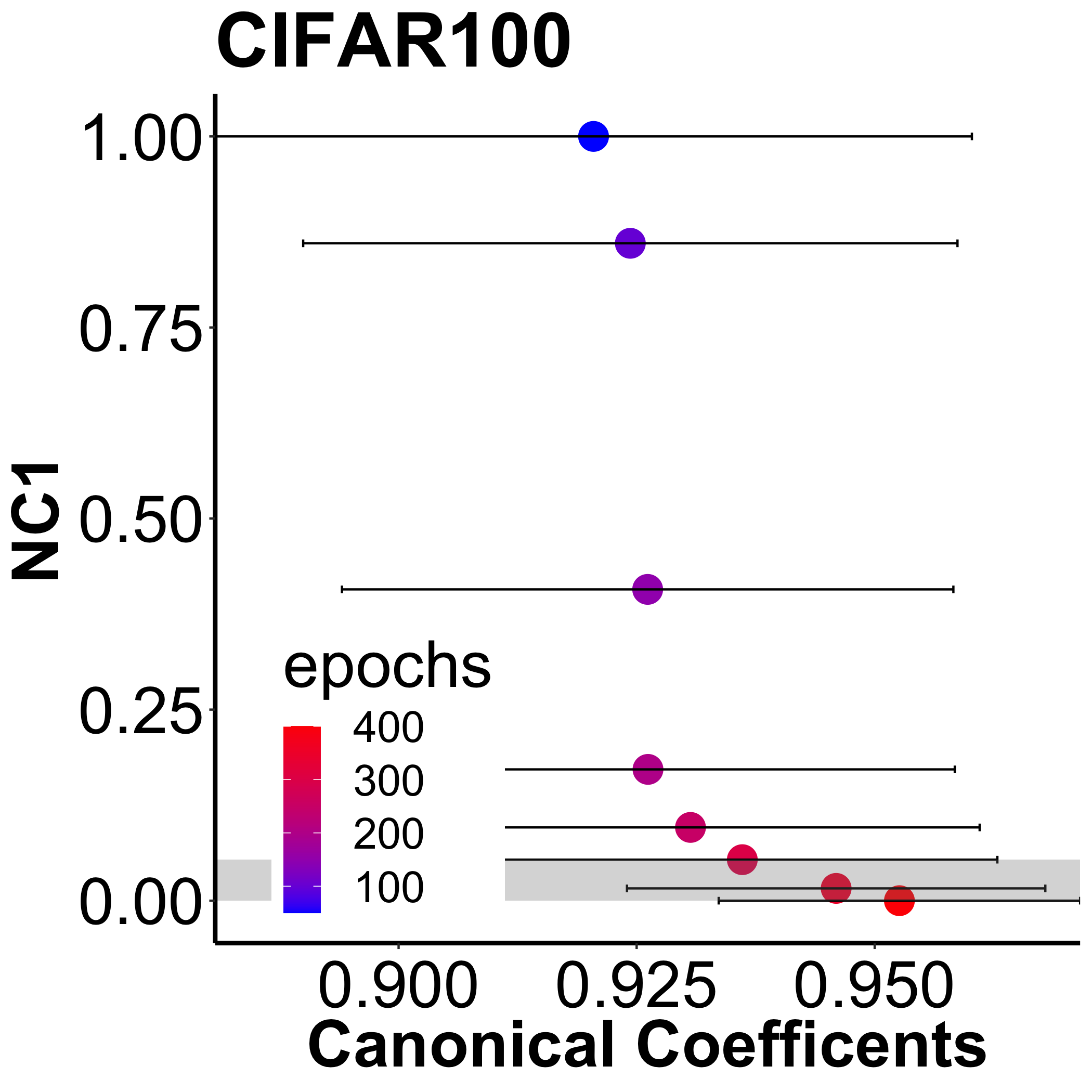}
    \label{fig:meanangle}
     \end{subfigure}
\caption{The emergence of neural collapse, shown via variability collapse (NC1), improves linear identifiability. We measure the linear identifiability between two learned representations using the average of their CCA coefficients \cite{Raghu2017}. Using CCA coefficients to measure linear similarity was proposed in \cite{Roeder2020}. X-axis shows the variance of $K$-leading CCA coefficients for $K$-class datasets (e.g., $K$=10 for CIFAR-10 and $K$=100 for CIFAR100). The Y-axis is the metric for variability collapse. Each dot is the mean of the CCA and the error bar shows the respective standard deviation. Colors are epochs. We also shade the area with less than 2\% training error (the terminal phase of training \cite{Ekambaram2017,Mueller2019}). a) CIFAR-10; b) CIFAR-100;}
\label{fig:NC}
\vspace{-15pt}
\end{figure}

We define neural collapse at the embedding layer (of a ResNet50 backbone) when we see the training accuracy stabilize around 98\%. This is because supervised contrastive learning works differently than cross entropy and MSE loss. At a training accuracy around 98\%, our test accuracy is already higher than those reported in \cite{Papyan2020, han2022neural} (e.g., for CIFAR100, our test accuracy is 76\% and it was 68\% in \cite{Papyan2020}). Meanwhile, previous work showed that popular datasets contain unexpected labeling inconsistencies \cite{Ekambaram2017,Jadari2019,Mueller2019}. We observe that linear identifiability between learned representations improves significantly following the emergence of neural collapse (shaded regions in Fig. \ref{fig:NC}). We did not observe a significant improvement in linear identifiability prior to neural collapse. After neural collapse, we find that the linear identifiability of the learned representation improves in terms of both higher means and a smaller standard deviation between relevant canonical coefficients. This may correlate with the emergence of the $K$-simplex ETF ($K$=10 for CIFAR10, 100 for CIFAR100, respectively). We evaluate this hypothesis in the next section. 

In addition, we observe that the linear classification performance of models trained with ImageNet32 stabilizes around 81\%, while their test performance is comparable to the state-of-the-art using a vision transformer \cite{Yang2022}. This behavior does not fit the definition of the terminal phase of training (training error goes to zero) originated from cross entropy trained models. However, it is consistent with our information theoretical interpretation of neural collapse, i.e., the model does not retain additional information about the label from its input. We interpret this result as an outcome of training an encoder with the supervised contrastive loss \cite{Oord2018, Khosla2020}, as opposed to training linear classification directly (linear classifiers are usually trained separately after the training of the encoder finishes). In Section \ref{section:collapse}, we use models pretrained with ImageNet-32 as feature extractors. In zero-shot transfer learning for CIFAR10, neural collapse in feature activations allows us to compress the representation for CIFAR10 to lower dimensions (much less than 2048). 

\vspace{-0.1in}
\subsection{$I(Z_1;Z_2)$ contains the majority of the Classification-relevant information}
An important advantage of linear identifiability is that all models may converge to equivalent solutions \cite{Hyvarinen2018, Roeder2020}. If this is the case, then the learned representation is stable in the sense that its performance on downstream tasks is consistent despite random initialization. This theoretical prediction is validated in Table \ref{table:test_main} by demonstrating that two models of the same architecture trained in parallel with a supervised contrastive learning objective have comparable generalization performance and make comparable decisions on the test dataset. This observation also suggests that the mutual information between the learned representations $Z_1$ and $Z_2$ contains the majority of the classification-relevant information. Next, we investigate the structure of the correlation between $Z_1$ and $Z_2$.
\begin{table}[t]
\vskip 0.15in
\begin{center}
\begin{small}
\begin{sc}
\begin{tabular}{lcccr}
\toprule
Data set & Cifar-10 & Cifar-100 & ImageNet32 \\
\midrule
model 1    & 95.4& 75.7& 56.8 \\
model 2 & 95.8& 76.0& 56.5\\
Both    & 93& 69.2& 56.4 \\
\bottomrule
\end{tabular}
\end{sc}
\end{small}
\end{center}
\caption{\textbf{Test performance shared by two deep neural networks trained in parallel}: We show here the test accuracy of either model 1 or model 2. We also show the fraction of the samples that are classified correctly by both models. Linearly identifiable representations from ResNet50 trained with CIFAR10, CIFAR100 and ImageNet32 learn similar decision boundaries for classifications. }
\vspace{-25pt}
\label{table:test_main}
\end{table}

\vspace{-0.1in}
\subsection{Supervised contrastive learning compresses classification into a $K$-dimensional representation as generalization improves}
\label{section:ETF}

\begin{table}[t]
\vskip 0.15in
\begin{center}
\begin{small}
\begin{sc}
\begin{tabular}{lcccr}
\toprule
Data set & CIFAR10 & CIFAR100 \\
\midrule
Raw (2048D)   & 95.9& 75.9 \\
Ranked (2048D) & 95.8& 75.6&\\
Ranked $K$-IB (10D or 100D)  &93.8  &  73.5\\
\bottomrule
\end{tabular}
\end{sc}
\end{small}
\end{center}
\caption{Test accuracy using $K$ dimensional IB optimal representations vs. using the raw 2048 D representation or its rank transformed version. $K$=10 for CIFAR10 and $K$=100 for CIFAR100. The similarity between the IB optimal representations and the 2048D representation suggests that supervised contrastive learning compresses classification relevant information into the $K$ dimensional IB optimal representation.}
\label{acc:IB}
\vspace{-25pt}
\end{table}

According to prior research \cite{Zhu2021,Graf2021}, learning a $K$-simplex ETF requires at least $K$ dimensions. Given that, following the emergence of neural collapse, clusters of different classes become non-overlapping in the high-dimensional encoder space, we question whether the improvement in linear identifiability brought by neural collapse concentrates on the first $K$ leading dimensions. The insight from Table \ref{table:test_main} also prompts us to question whether the correlation structure within $I(Z_1;Z_2)$ becomes more linearly similar. In addition, we also observed the CCA coefficients between the raw learned representation and its rank-transformed version, after 400 epochs of training (see Supplementary Information).

At the 400th epoch, we observed high performance on both training and testing datasets for CIFAR10 and CIFAR100 ( Table \ref{table:test_main} and Supplementary Information). Therefore, we hypothesize that this is the stage where the models approximate the entropy coding of $H(Y) -\delta$ ($\delta$ is a small error constant). This is an optimal solution of information bottleneck (IB) for classification. In general, the IB objective is intractable. However, the linear similarity enables us to construct a Meta-Gaussian information bottleneck (MGIB) between two learned representations. We hypothesize that the close similarity between the rank transformed data and the raw data suggests that the optimal solution to the MGIB may capture the statistical structure within the $K$-dimensional subspace for the emerging $K$-simplex ETF ($K=10$ for CIFAR10, $K=100$ for CIFAR100). In Table \ref{acc:IB}, we show that the classification performance using the MGIB-optimal representation, which is only $K$-dimensional, retains nearly all the classification performance achievable from the raw, 2048 dimensional learned representation. We also observed that as training progresses, supervised contrastive learning compresses most of the information relevant to classification into the IB optimal $K$-dimensional representations while improving test generalization (see Supplementary Information).


\vspace{-0.1in}
\subsection{$K$-simplex ETF emerges within the IB optimal K-dimensional representations as training progresses}
\label{section:collapse}
\begin{figure}
     \centering
     \begin{subfigure}[b]{0.24\textwidth}
     \caption{}
         \centering
         \includegraphics[width=0.9\textwidth]{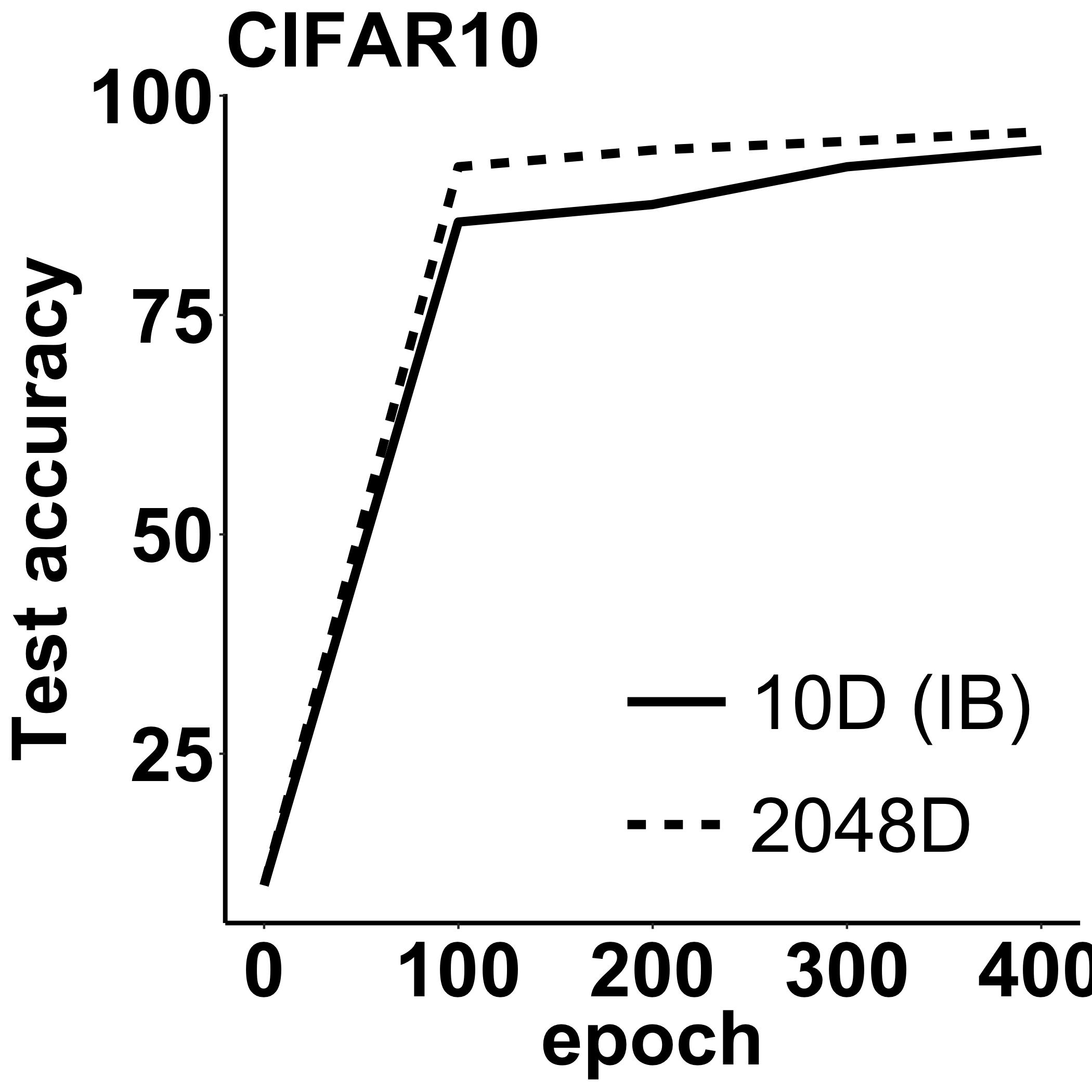}
     \label{fig:accuracy_cifar10}
\end{subfigure}
 \begin{subfigure}[b]{0.24\textwidth}
     \caption{}
         \centering
         \includegraphics[width=0.9\textwidth]{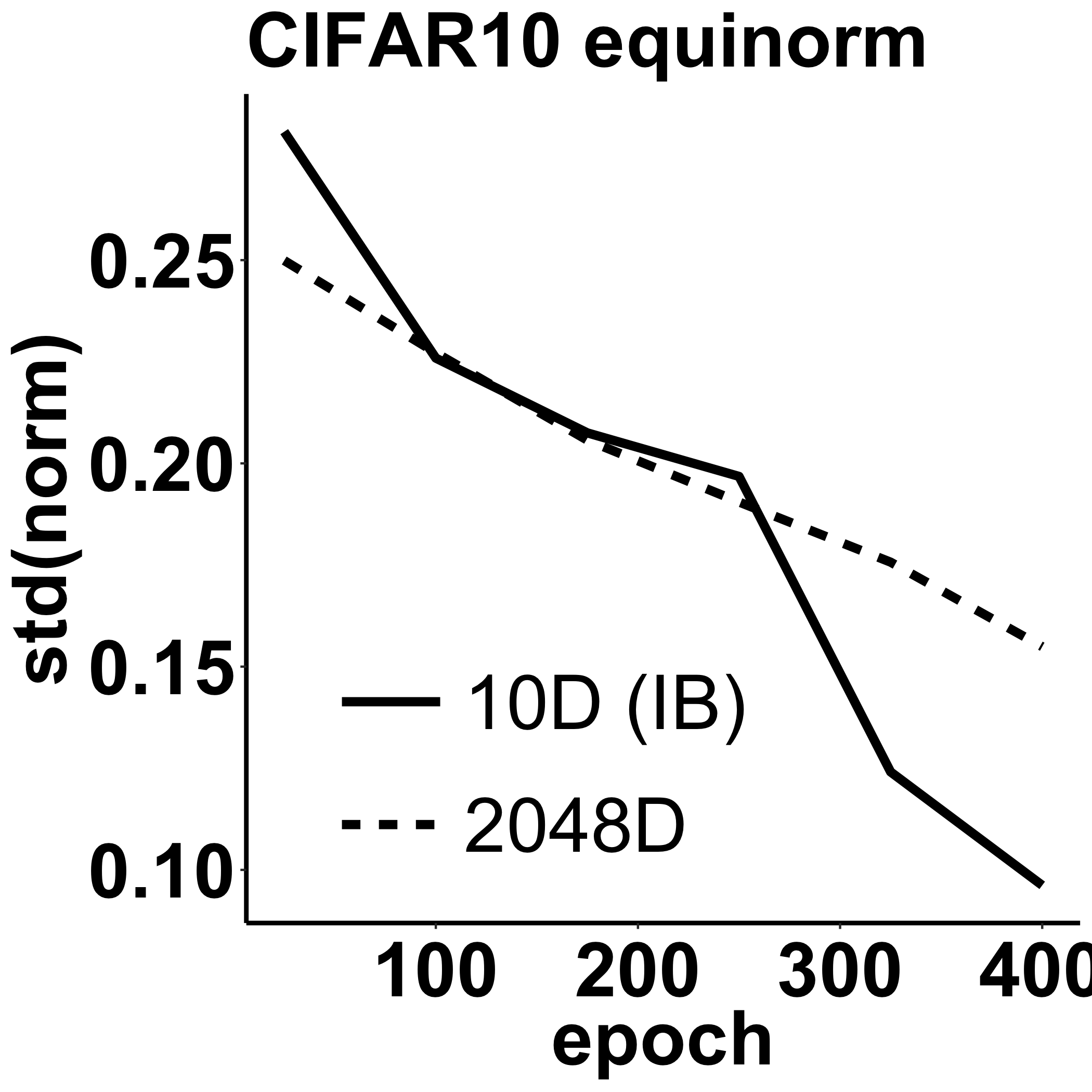}
     \label{fig:norm_cifar10}
\end{subfigure}
 \begin{subfigure}[b]{0.24\textwidth}
     \caption{}
         \centering
         \includegraphics[width=0.9\textwidth]{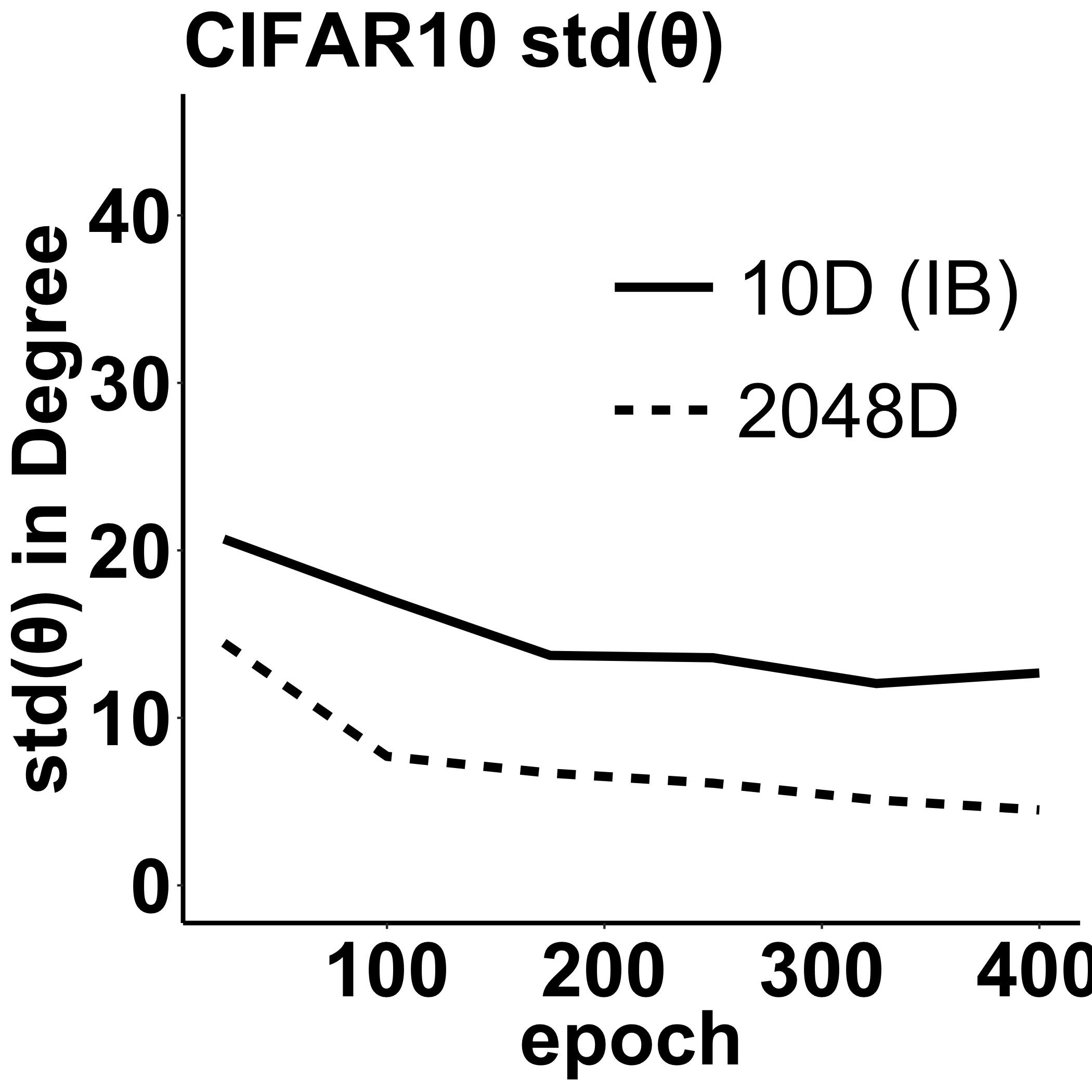}
     \label{fig:std_cifar10}
\end{subfigure}
 \begin{subfigure}[b]{0.24\textwidth}
     \caption{}
         \centering
         \includegraphics[width=0.9\textwidth]{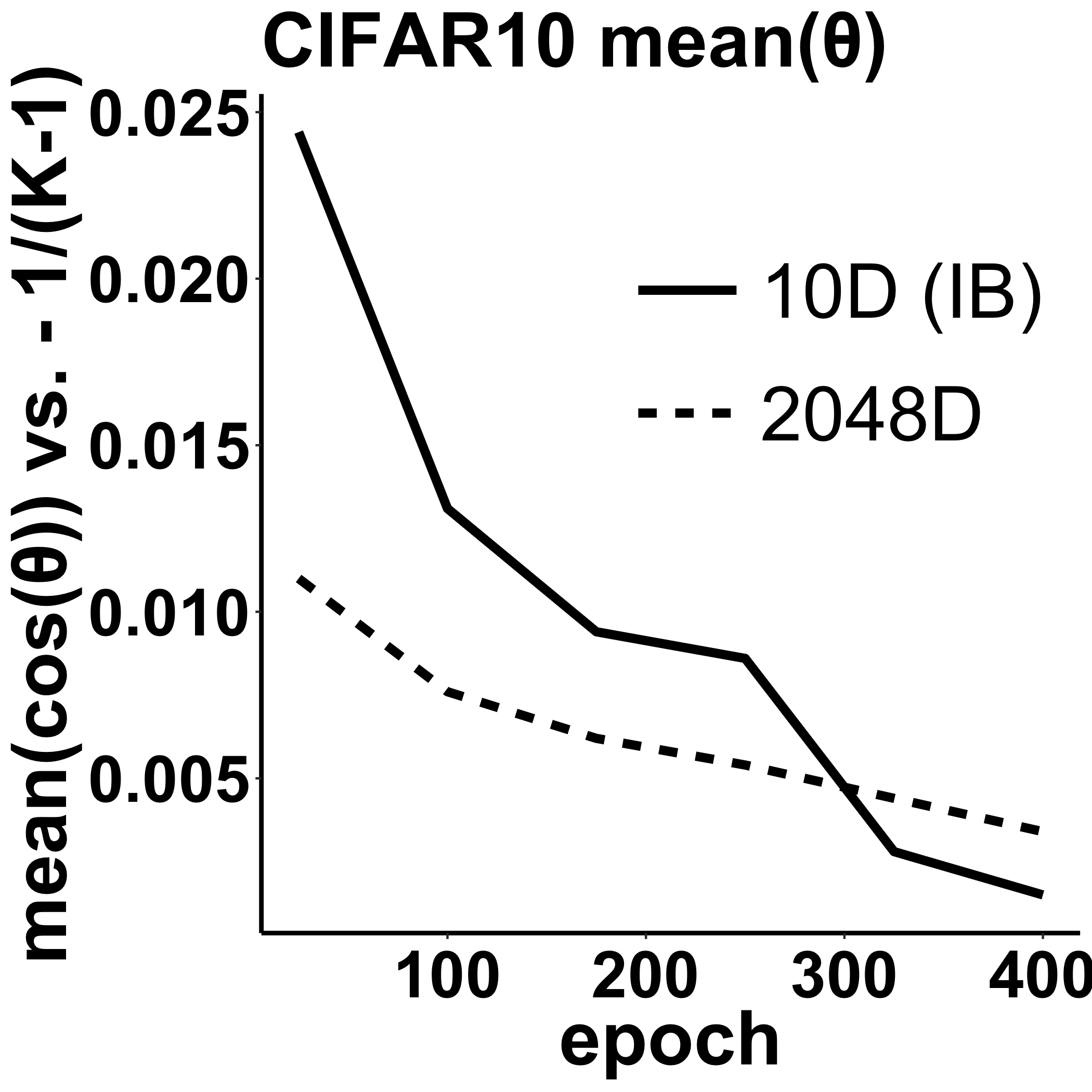}
     \label{fig:mean_cifar10}
\end{subfigure}\\
     \begin{subfigure}[b]{0.24\textwidth}
     \caption{}
         \centering
         \includegraphics[width=0.9\textwidth]{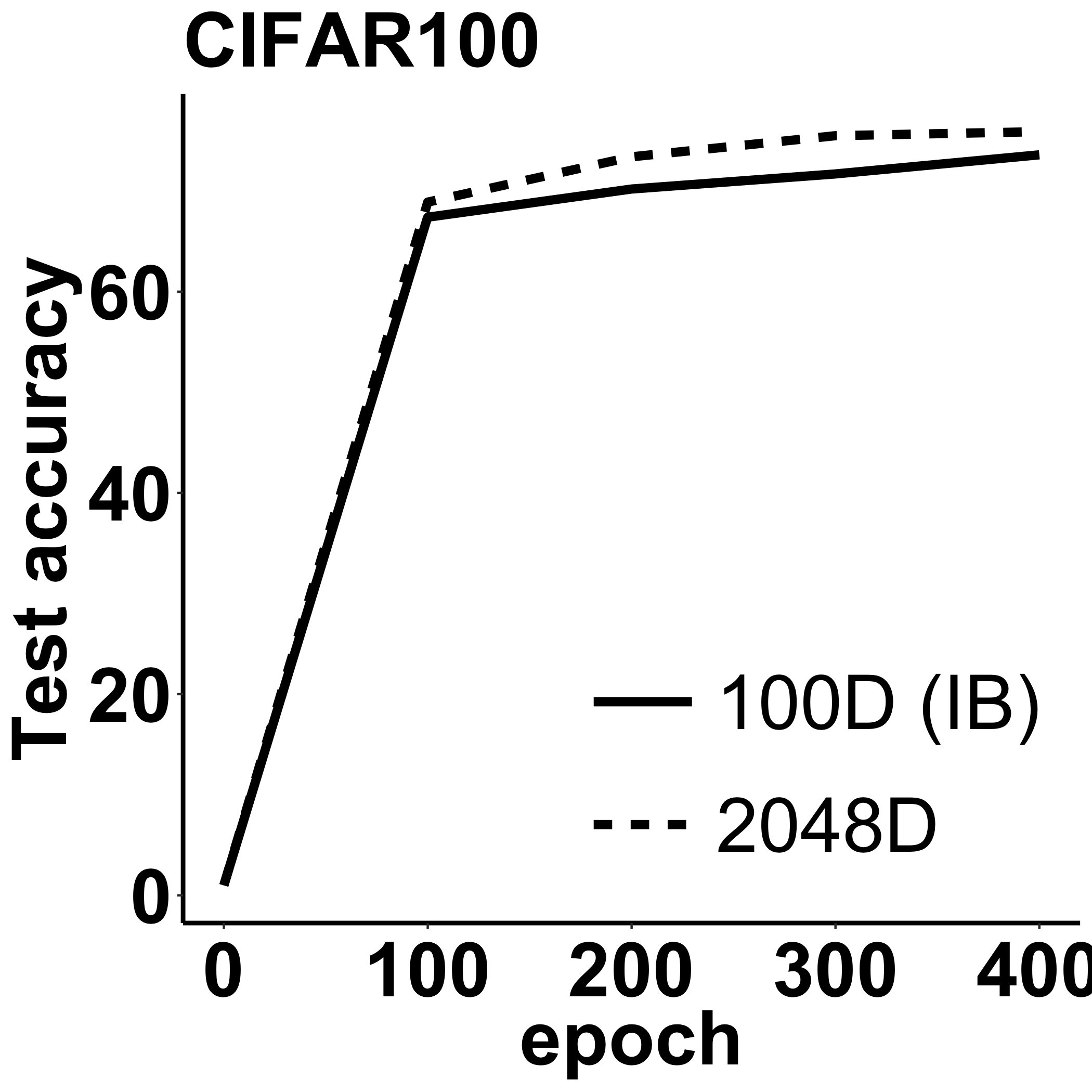}
    \label{fig:accuracy_cifar100}
     \end{subfigure}
      \begin{subfigure}[b]{0.24\textwidth}
     \caption{}
         \centering
         \includegraphics[width=0.9\textwidth]{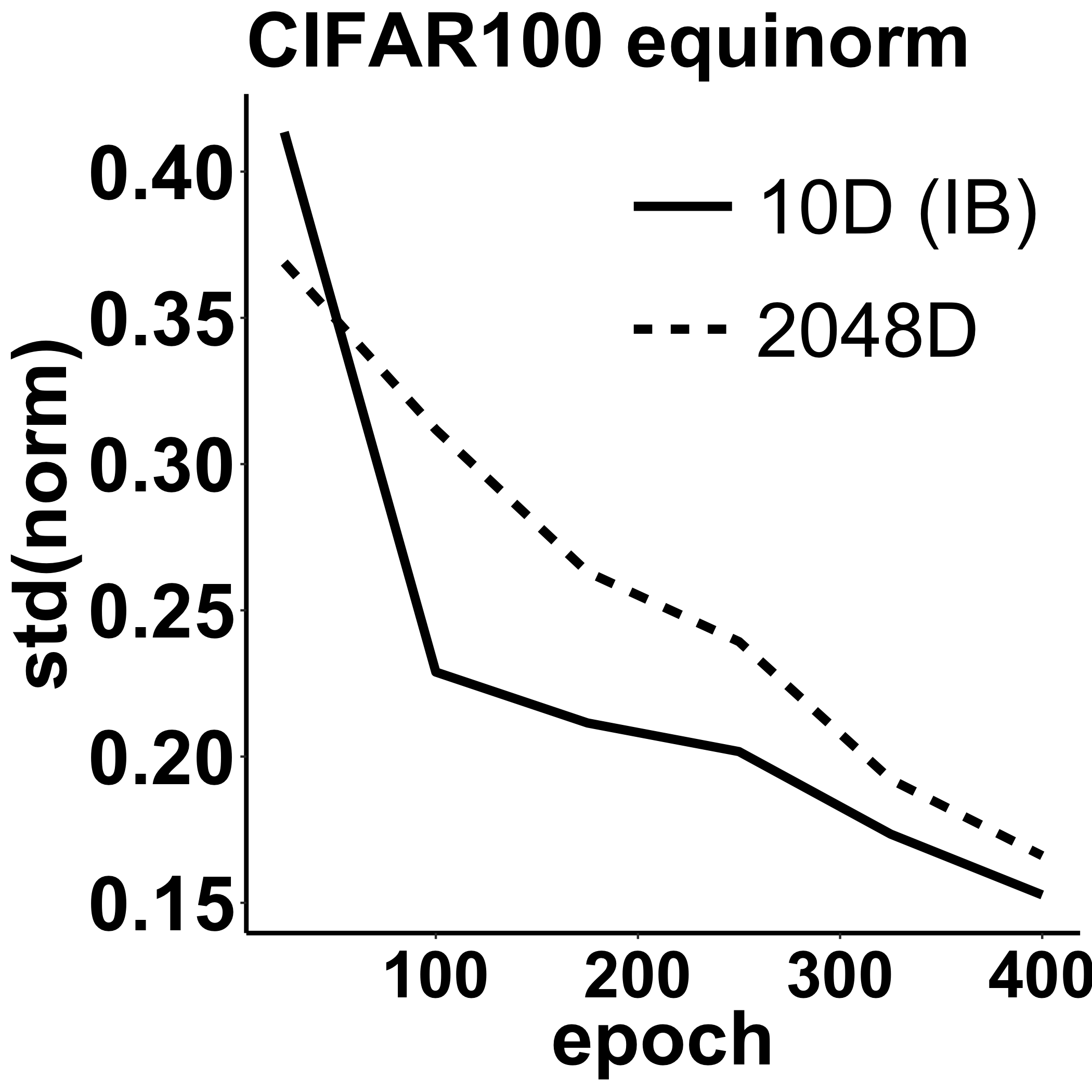}
    \label{fig:norm_cifar100}
     \end{subfigure}
      \begin{subfigure}[b]{0.24\textwidth}
     \caption{}
         \centering
         \includegraphics[width=0.9\textwidth]{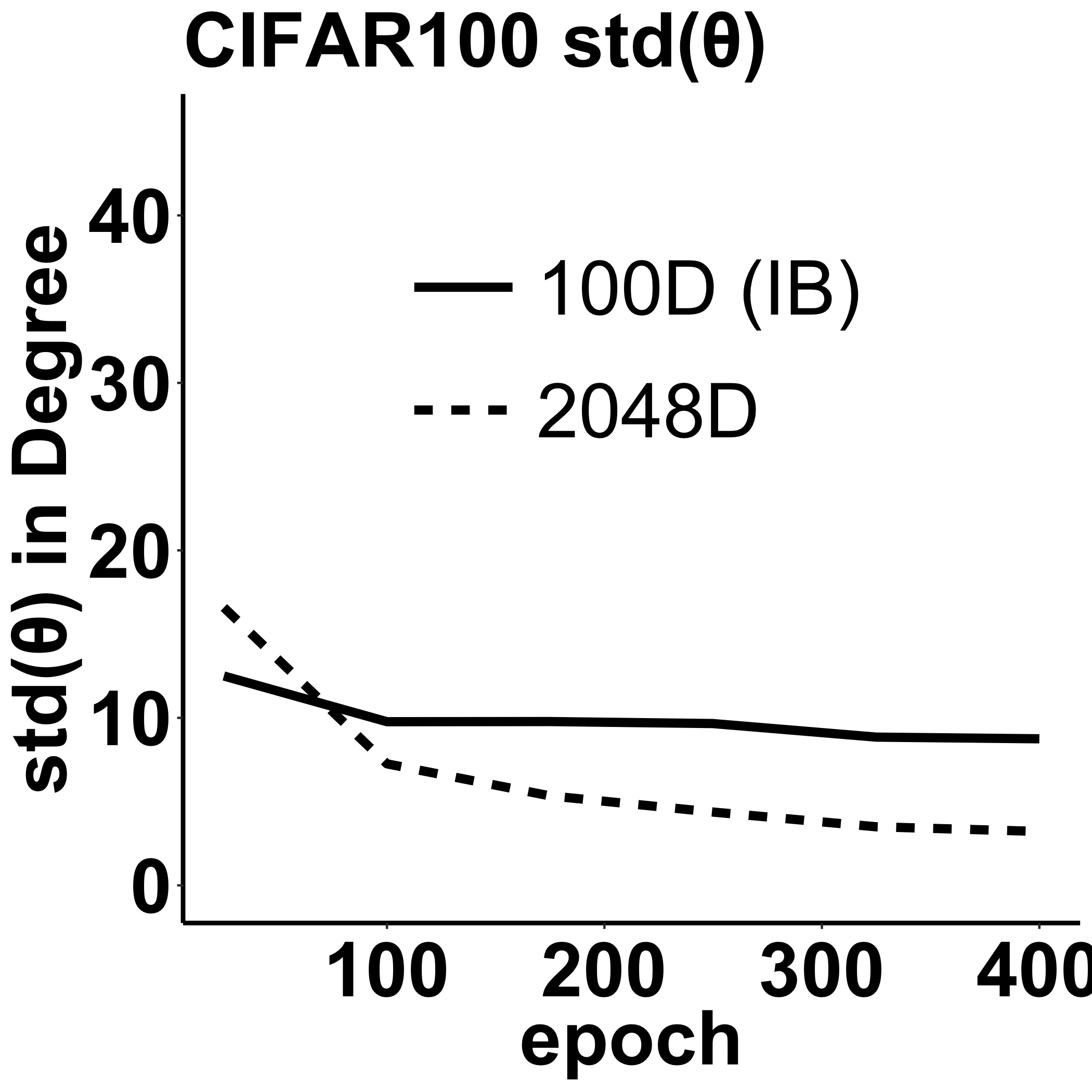}
    \label{fig:mean_cifar100}
     \end{subfigure}
      \begin{subfigure}[b]{0.24\textwidth}
     \caption{}
         \centering
         \includegraphics[width=0.9\textwidth]{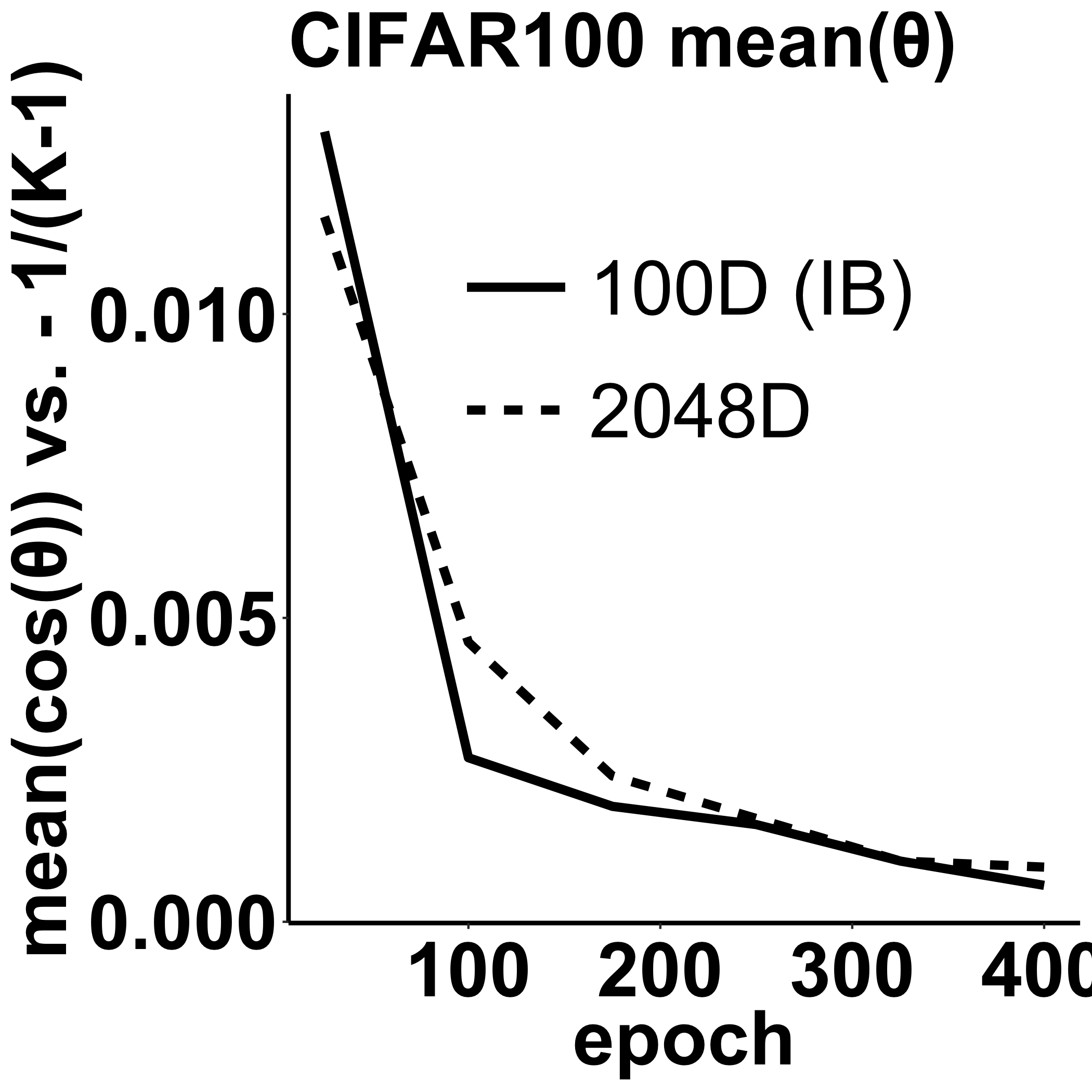}
    \label{fig:std_cifar100}
     \end{subfigure}
\caption{Within the $K$-dimensional IB optimal Gaussian distribution for both CIFAR10 and CIFAR100, we observe that class means exhibit K-simplex ETF comparable to the K-simplex ETF in the full 2048-D representation space. A) As training progresses, the $K$ dimensional IB optimal representation retains most of the classification performance for CIFAR10. B) As training progresses, the standard deviation of norm for all class means, i.e., $Std_k(||\mu_k-\mu_{all}||_2/Avg(||\mu_k-\mu_{all}||_2))$ gets smaller within the $K$-dimensional Gaussian distribution (shown in solid line). The diminishing effect is more salient than those observed in the 2048-D full representation space (shown in dashed line). C) The standard deviation of angles between class means, i.e., $\theta_{\mu_{c,c'}}$ for CIFAR10 gets smaller as training progresses. The magnitude of the difference between $K$-dimensional IB optimal Gaussian distribution (shown in solid line) and the full 2048-D representation space (shown in dashed line) is small, i.e., $5^\circ$. D) The cosine of mean angles between class activations converge to $(-1/(K-1))$ (K=10 for CIFAR10), i.e. $\cos{\theta_{\mu_{c,c'}}}\rightarrow (-1/(K-1))$. Again, the $K$-dimensional IB optimal Gaussian distribution behaves comparable to the full 2048-D representation space in placing different class means to pairwise angles close to $\cos^{-1}{(-1/(K-1))}$. E)-H) K-simplex ETF observations for CIFAR100. }
\label{fig:IBNC4}
\vspace{-20pt}
\end{figure}
We first show that ResNet50 trained with supervised contrastive loss compresses most of the information relevant to classification into the IB optimal $K$-dimensional representations while improving generalization (shown as test accuracy in Fig. \ref{fig:accuracy_cifar10} for CIFAR10 and Fig. \ref{fig:accuracy_cifar100} for CIFAR100). This ties compression with improved test generalization (see Supplementary Information for further analysis). This motivates us to investigate whether $K$-simplex ETF also emerges in the compressed MGIB-optimal representation (i.e., optimal solutions for $I(Z_1;T)-\beta I(T;Z_2)$). Following \cite{Papyan2020}, we also show whether the class clusters converge to an equal norm and whether the angles among them converge to $\cos^{-1}(-1/(K-1))$. Fig. \ref{fig:norm_cifar10} and \ref{fig:norm_cifar100} show that the variation in the norms of the class means decreases. We find that the class means within the MGIB-optimal representation converge to an equinormed state within the compressed MGIB-optimal representation (solid line). This convergence is comparable to that observed using the 2048D full representation space. Similarly, Fig. \ref{fig:std_cifar10} and \ref{fig:std_cifar100} indicate that all pairs of class means approximate equal sized angles through training. The difference in the magnitude of standard deviations between the compressed MGIB-optimal representation and the 2048D full representation space is small, i.e., $5^\circ$ to $7^\circ$. Fig. \ref{fig:mean_cifar10} and \ref{fig:mean_cifar100} further show that the angles between class means converge to the maximal possible angle, i.e., $\cos^{-1}(-1/(K-1))$ given the number of classes. We also observed the emergence of NC1 and NC4 within the compressed MGIB-optimal representation (see Supplementary Information). All these observations combined together imply that $K$-simplex ETF also emerges within the compressed MGIB-optimal representation as a signature of neural collapse. This respective $K$-simplex ETF is an optimal feature of source coding. Note that this is different from the previous information theoretic perspective which established the optimality of $K$-simplex ETF for linear classification.

\vspace{-0.1in}
\subsection{ImageNet32 trained models use noisy $K'$-dimensional linear representations in zero-shot transfer learning}
\label{section:imagenet}

\begin{figure}
     \centering
      \begin{subfigure}[b]{0.3\textwidth}
     \caption{}
         \centering
         \includegraphics[width=0.9\textwidth]{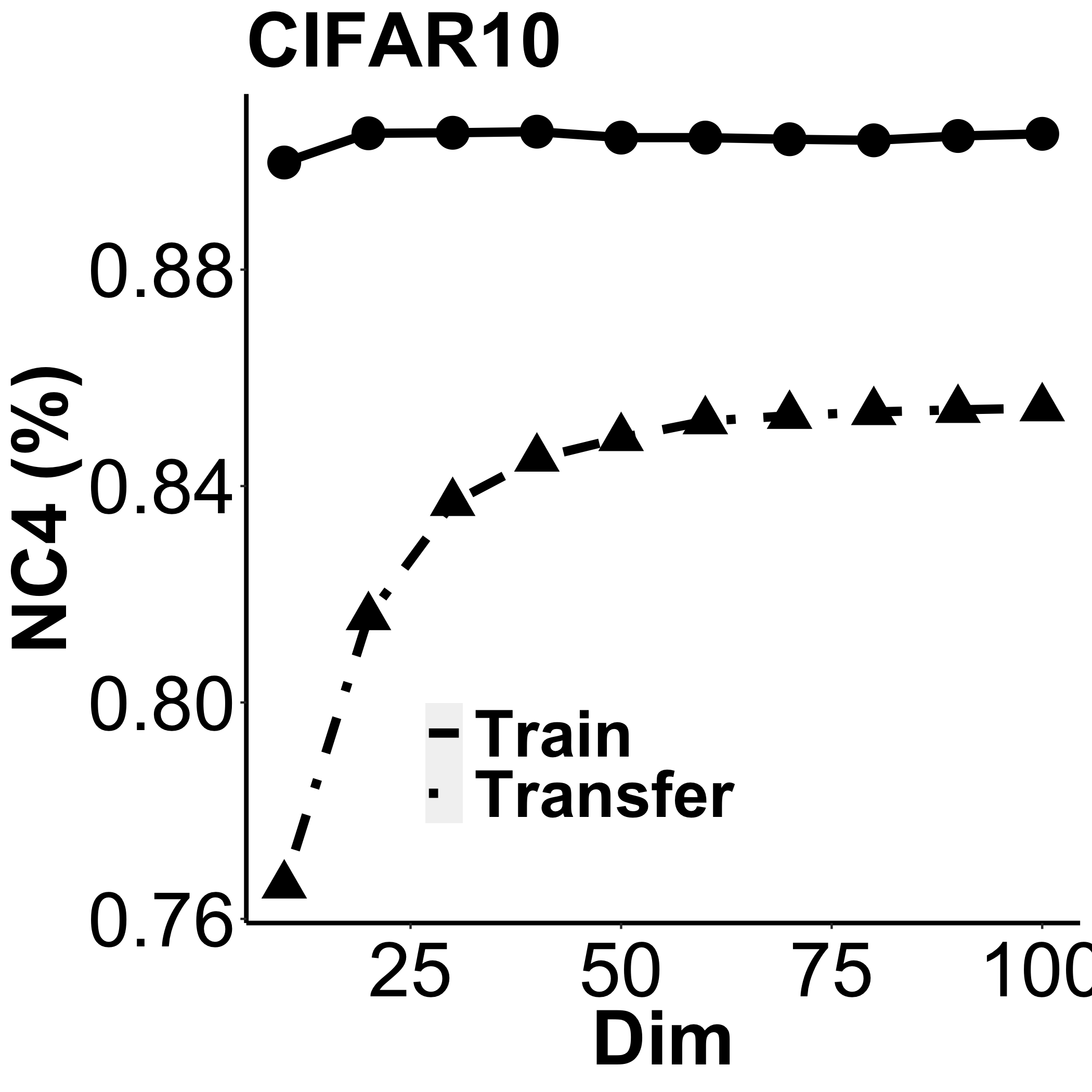}
     \label{fig:IBNC4_cifar10_transfer}
\end{subfigure}
     \begin{subfigure}[b]{0.3\textwidth}
     \caption{}
         \centering
         \includegraphics[width=0.9\textwidth]{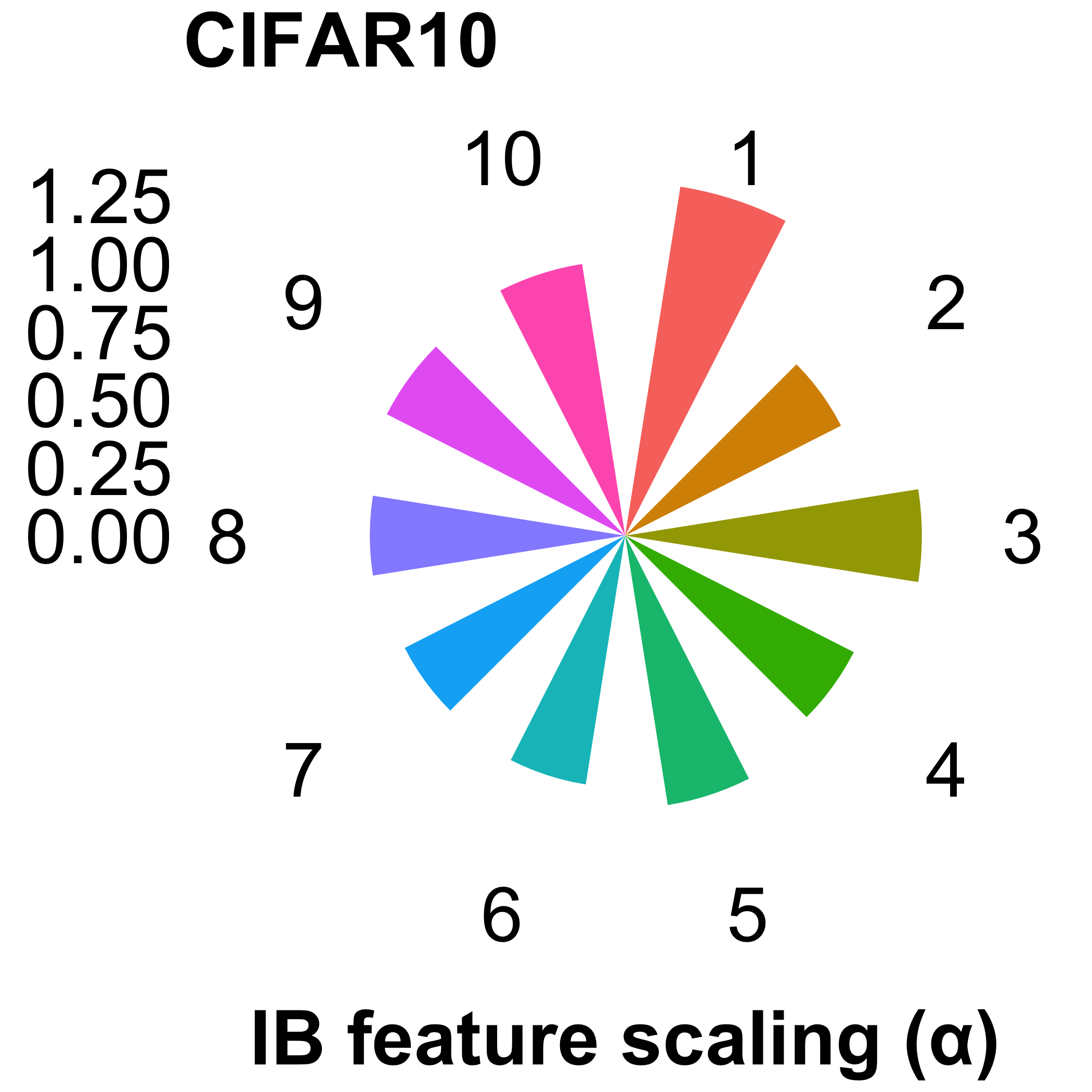}
     \label{fig:IBalpha_cifar10}
\end{subfigure}
     \begin{subfigure}[b]{0.3\textwidth}
     \caption{}
         \centering
         \includegraphics[width=0.9\textwidth]{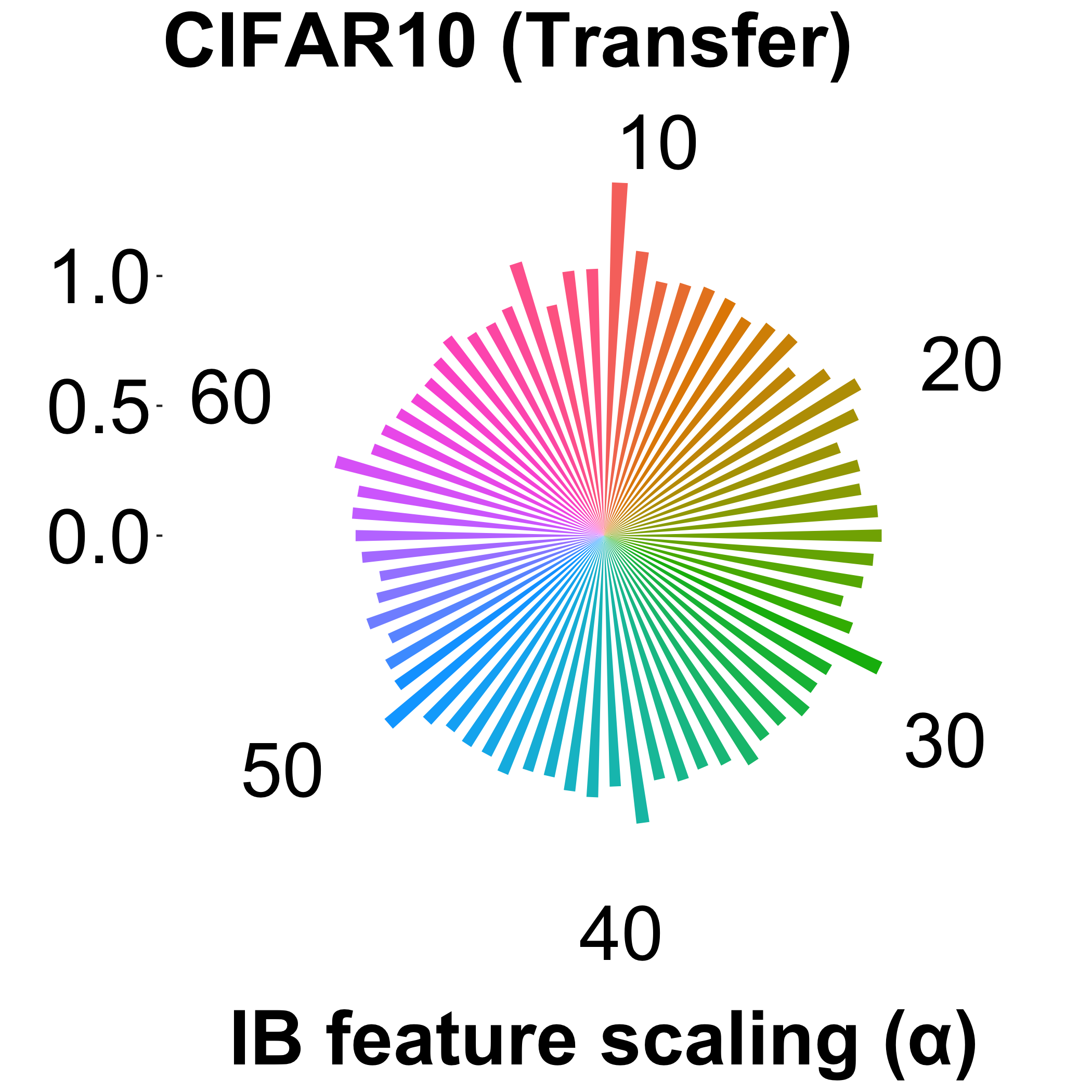}
    \label{fig:IBNC_cifar10_zeroshot}
     \end{subfigure}
\caption{A) The zero-shot transfer learning for CIFAR10 needs nearly 70 dimensions to enable classifiers to search for nearest class means (NC4) and retain classification performance. B) Feature scaling for the $K$-dimensional Gaussian distribution that fits the corresponding $K$-simplex ETF optimal for CIFAR10. Most of the feature scaling are between 0.9 and 1.1 with respect to the mean. Similar scalings between features indicates that each IB dimension corresponds to similar phase transition coefficients and contributes nearly equally to encode $H(Y)-\delta$. This is compatible with the geometry of $K$-simplex ETF. C) In the zero-shot transfer learning scenario, we may find a $K'$-dimensional Gaussian distribution that retains most of the classification information for CIFAR10. While $K' > K$ ($K' = 70$ and $K=10$ for CIFAR10), all feature scalings are similar (also between 0.9 and 1.1), indicating that the each IB dimension contributes near equivalently. Note that this suggests models trained with different datasets characterizes the same CIFAR10 with a similar geometry (i.e., the simplex ETF). Whether such geometry is a universal solution for reprsentation learning is an interesting future direction.}
\label{fig:IBzeroshot}
\end{figure}

We find that the models pretrained with ImageNet32 do not exhibit neural collapse on the 1000 ImageNet classes. To our surprise, simplex ETF still occurs when we use these models to perform zero-shot transfer learning for CIFAR10 (see Supplementary Information for additional results). In zero-shot transfer learning, the pretrained model obtains representations for $K'$-simplex ETF ($K' =70$ for CIFAR10; This is significantly less than 1000 but more than $K=10$ for CIFAR10) to retain 93.2\% classification performance. The emergence of this $K'$-simplex ETF simplifies classification as finding the nearest class mean as well (shown in Fig. \ref{fig:IBNC4_cifar10_transfer}). We next show that the $K'$-dimension Gaussian distribution for zero-shot transfer learning has a similar geometry compared to the $K$-dimension Gaussian distribution we obtained from models trained for CIFAR10. The magnitude of dimensions shown in Fig. \ref{fig:IBalpha_cifar10} suggests that all of the 10 IB optimal dimensions for the $K$-dimension Gaussian distribution have similar scaling. This indicates that these IB dimensions contribute to classification (or retaining information about $Y$) near-equivalently. Combining Fig.\ref{fig:IBalpha_cifar10} and \ref{fig:IBNC_cifar10_zeroshot}, we observe that models trained with different datasets converge to use simplex ETF to represent classification-relevant information for CIFAR10. We also observe a similar geometry when we compress subsets of ImageNet with far fewer classes (see Supplementary Information). 

\begin{table}[t]
\begin{center}
\begin{tabular}{lcccr}
\toprule
Representation&Raw (2048D)&Ranked (2048D)&IB(70D)\\
\midrule
CIFAR10&95.8\%&93.9\%&93.2\%\\
\bottomrule
\end{tabular}
\caption{Zero-shot transfer learning (CIFAR10) with models pretrained using ImageNet-32. The 70D-IB compresses the majority of compressible classification relevant information. We also observe a similar compression for subsets of ImageNet-32 with fewer classes (see Supplementary Information).}
\label{acc:transfer}
\end{center}
\vspace{-20pt}
\end{table}

\vspace{-0.1in}
\section{Conclusion}
\vspace{-0.05in}
Overall, we observe that representations obtained by supervised contrastive learning contain unique geometries when neural collapse and linear identifiability are combined. Supervised contrastive learning models approximate a special optimal solution for information bottleneck problem of classification due to neural collapse, and linear identifiability allows us to interrogate the geometries of this optimal solution. These geometries allow pairs of learned representations to be correlated through a low dimensional Gaussian copula for small datasets (with a small number of classes $K$). Due to the equiangularity and maximum norm of the $K$-simplex ETF, such a Gaussian copula has nearly equivalent covariance across all of its dimensions. Its low-dimensionality improves its generalization \cite{Shamir_2010}. Although we do not observe convergence to the $K$-simplex ETF with models trained using ImageNet32, we observe that they exhibit a noisier version of neural collapse during zero-shot transfer learning with CIFAR10 and small subsets (see Supplementary Information). This demonstrates that, despite the fact that the training dynamics for small and large datasets may differ in contrastive learning, our insight that neural collapse in supervised contrastive learning may lead to better generalization remains the same. Therefore, this finding may apply to a large number of models with contextualized data representations. For example, since modeling low-resource languages often requires fine-tuning from pretrained large language models \cite{Griesshaber2020, Mahabadi2021}, it will be interesting to determine whether finding a simplex ETF representation leads to effective fine-tuning.

\section*{Acknowledgement}
This work was supported by the National Science Foundation through the Physics Frontier Center for Living Systems (PHY-2317138) and the NSF-Simons National Institute for Theory and Mathematics in Biology, awards NSF DMS-2235451 and Simons Foundation MP-TMPS-00005320. 
\bibliographystyle{plainnat}
\bibliography{contrastive}
\end{document}